\documentclass[sigconf]{acmart}
\AtBeginDocument{%
  \providecommand\BibTeX{{%
    \normalfont B\kern-0.5em{\scshape i\kern-0.25em b}\kern-0.8em\TeX}}}
    
\usepackage{amsmath}
\usepackage{subcaption}
\usepackage{bm}
\usepackage{algorithm}
\usepackage[noend]{algpseudocode}
\usepackage[utf8]{inputenc}

\setcopyright{rightsretained}
\copyrightyear{2020} 
\acmYear{2020} 
\acmDOI{10.1145/3371382.3378286}

\acmConference[HRI '20 Companion]{Companion of the 2020 ACM/IEEE International Conference on Human-Robot Interaction}{March 23--26,2020}{Cambridge, United Kingdom}
\acmBooktitle{Companion of the 2020 ACM/IEEE International Conference
	on Human-Robot Interaction (HRI '20 Companion), March 23--26, 2020,Cambridge, United Kingdom}
\acmDOI{10.1145/3371382.3378286}
\acmISBN{978-1-4503-7057-8/20/03}

\begin{document}

\title[]{Learning, Generating and Adapting Wave Gestures for Expressive Human-Robot Interaction}

\author{Mihalis Panteris}
\affiliation{%
	\institution{MINES ParisTech}
	\streetaddress{60 Boulevard Saint-Michel}
	\city{Paris}
	\country{France}
	\postcode{75272}
}
\email{mihalis.panteris@mines-paristech.fr}
\author{Simon Manschitz}
\affiliation{%
	\institution{Honda Research Institute Europe}
	\streetaddress{Carl-Legien-Straße 30}
	\postcode{63073}
	\city{Offenbach/Main}
	\country{Germany}
}
\email{simon.manschitz@honda-ri.de}
\author{Sylvain Calinon}
\affiliation{%
	\institution{Idiap Research Institute}
	\streetaddress{Rue Marconi 19}
	\postcode{1920}
	\city{Martigny}
	\country{Switzerland}
}
\email{sylvain.calinon@idiap.ch}

\begin{abstract}
	This study proposes a novel imitation learning approach for the stochastic generation of human-like rhythmic wave gestures and their modulation for effective non-verbal communication through a probabilistic formulation using joint angle data from human demonstrations. This is achieved by learning and modulating the overall expression characteristics of the gesture (e.g., arm posture, waving frequency and amplitude) in the frequency domain. The method was evaluated on simulated robot experiments involving a robot with a manipulator of 6 degrees of freedom. The results show that the method provides efficient encoding and modulation of rhythmic movements and ensures variability in their execution.
\end{abstract}
\maketitle

\section{Introduction}
Variability is in the very core of human motor behaviour \cite{Stergiou2006}. While many humanoid robots can perform basic wave gestures, these gestures are usually hard-coded behaviours. Consequently, the gesture looks rather stiff since there is no variance in the execution of the movement. Methods such as probabilistic movement primitives (ProMPs) \cite{Paraschos2018} encode the variability of a movement from a set of demonstrations through a probabilistic trajectory representation. Additionally, ProMPs take advantage of probability distribution properties in order to preserve this variability when adapting the movement to different final positions or via-points. Despite this being straightforward for stroke-based movements, preserving the stochasticity and the variability in the execution of the movement is not possible when modulating rhythmic movements such as wave gestures. In this study we propose that learning should take place in the frequency domain in order to overcome these issues and to efficiently exploit demonstrations of diverse motor patterns. In \cite{Calinon19MM}, it was shown that the basis functions used in probabilistic movement primitives do not need to be restricted to radial bases, and can take other forms such as Fourier basis functions. The proposed framework exploits the conditioning property of normal distributions for modulating the gesture. 

\section{The proposed learning framework}

\begin{algorithm}
	\caption{Learning Algorithm}
	\label{FMP}
	\begin{algorithmic}[1]
		\State \textbf{Data:} a set of $M$ trajectories $\textbf{y}$ of time length T and dimension D in joint angle space.
		\State \textbf{Input:} number of conjugate basis functions K
		%\State \textbf{Result:} The mean $\bm{\mu_x}$ and covariance $\bm{\Sigma_x}$ of $\ln{\bm{w}}=\ln{\bm{w^r}}+i\,\bm{w^\theta}$
		%$\ln{\bm{W}}=\ln{ \left| w \right|}+i \text{Arg}(w)$
		%\State Detect shortest data length among the trajectories in the set and discretize evenly all the demonstrations 
		\For{each trajectory m}
		%\State Compute the signal's base frequency $\omega_0=\frac{2\pi}{T}$
		\State Generate basis functions: $\bm{\Psi} = f(t,K,T,D)$
		\State Compute the weight vector $\bm{w_m}$:
		\begin{equation*}
		\label{complex_polar}    
		\begin{split}
		\bm{w_m} &= (\bm{\Psi}^H \bm{\Psi})^{-1}\bm{\Psi}^H \bm{y_m}  \\
		\bm{w^r_m} &= \left|\bm{w_m}\right|\\
		\bm{w^\theta_m} &= \text{Arg}(\bm{w_m})
		\end{split}
		\end{equation*}
		\EndFor
		\State Fit a Gaussian over the random vector $\bm{X}=\left[\ln{\bm{w^r}},\bm{w^\theta} \right]^\top$
		\begin{equation*}
		\bm{\mu_x} = \frac{1}{M} \sum_{m=1}^{M}\bm{x}_m, \quad 
		\bm{\Sigma_x} = \frac{1}{M} \sum_{m=1}^{M}(\bm{x}_m-\bm{\mu_x})(\bm{x}_m-\bm{\mu_x})^\top
		\end{equation*}
		\State \textbf{return} $\bm{\mu_x}$, $\bm{\Sigma_x}$
	\end{algorithmic}
\end{algorithm}

We propose here a framework for learning rhythmic movements in joint angle space similar to the one in \cite{Paraschos2018} for stroke-based movements. An overview of the learning method is presented in Algorithm 1. Inspired by Fourier analysis we deploy complex exponential basis functions in the context of movement primitives in order to approximate a trajectory $\bm{y_{m}} $ of time length T with 2*K conjugate basis functions plus an offset. In vector form this is expressed as:
\begin{equation}
\bm{y_m} = exp(i\,\dfrac{2\pi \bm{t} \bm{k}^\top}{T})\,\bm{w_m}=\bm{\Phi}\,\bm{w_m}
\label{one}
\end{equation}
With $t = \left[0,1,...,T\right]^\top, k=\left[-K,-K+1,...,K-1,K\right]^\top \text{and } \bm{\Phi} \in {\mathbb R}^{T\times(2K+1)}$. The above formulation is extended to multidimensional trajectories by using $\bm{\Psi}=\bm{\Phi} \otimes \bm{I}$, where $\bm{I}$ is the identity matrix of size D and $ \otimes $ is the Kronecker product. Thus the solution is found by the least squares estimate for complex valued variables. Since the weights $ \bm{w_m} $ are complex numbers they can be decomposed to their modulus $ \bm{w^r} $ and phase $ \bm{w^\theta} $ components. The representation of the complex weights in polar coordinates enables us to modulate intuitively the movement in terms of frequency, amplitude or phase. 

The modulus is a non-negative number and thus must be bounded in zero. By assuming that the least squares estimate of the weights yields non-zero values we can compute the principal value of the logarithm of $\bm{w}$:  
\begin{equation*}
\ln{\bm{w}}=\ln{\bm{w^r}}+i\,\bm{w^\theta},
\end{equation*}
where $\bm{w^\theta} \in \left(-\pi,\pi\right]  $, and fit a joint normal distribution over its real and imaginary parts, i.e fit a normal distribution over the random vector $\bm{X}=\left[\ln{\bm{w^r}}, \bm{w^\theta}\right]^\top $.
Through this distribution we can stochastically generate wave gestures by sampling over $\mathcal{N}(\mu_x,\,\Sigma_x)$ and then synthesize the resulting trajectory according to \eqref{one}: 
\begin{equation}
\bm{y_s}=\bm{\Psi} \exp\left(\ln{\bm{w^r}}\right) \exp\left(i\,\bm{w^\theta}\right)
\label{two}
\end{equation}

When the offsets $\bm{w^0}$ of the demonstrations have both negative and positive sign, a normal distribution should be fitted on the random vector  $\bm{X}=\left[\ln{\bm{w^r}}, \bm{w^\theta}, \bm{w^0}\right]^\top$ instead.

Similarly to \cite{Paraschos2018} where the conditional probability property is exploited to modulate the trajectory in joint angle space, we modulate the wave gesture in the weight space by conditioning on a subset of the random vector $\bm{X}$. Contrary to ProMPs a trajectory distribution can not be defined since $\bm{y}$ is not a linear transformation of $\bm{X}$.

\section{Experimental results}

\begin{figure}[h]
	\centering
	\includegraphics[width=\linewidth]{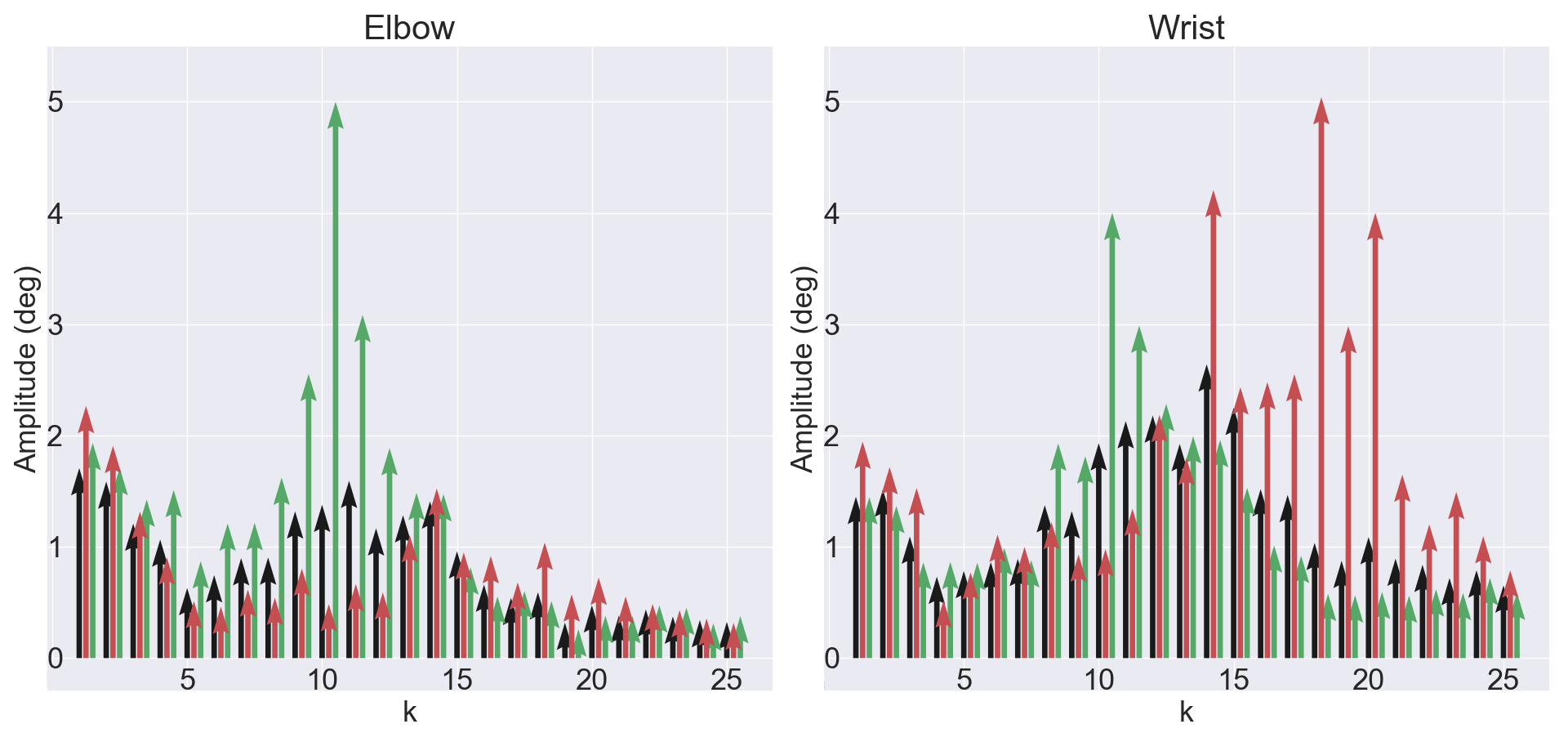}
	\caption{Mean of the single sided spectrum of the demonstrations  ($\mu_{\bm{w}^r}$) in black. Frequency modulation of the elbow joint in green. Frequency modulation of the wrist joint in red.}
	\Description{The 1907 Franklin Model D roadster.}
	\label{fig:spectrum}
\end{figure}

\begin{figure}[h]
	\centering
	\includegraphics[width=\linewidth]{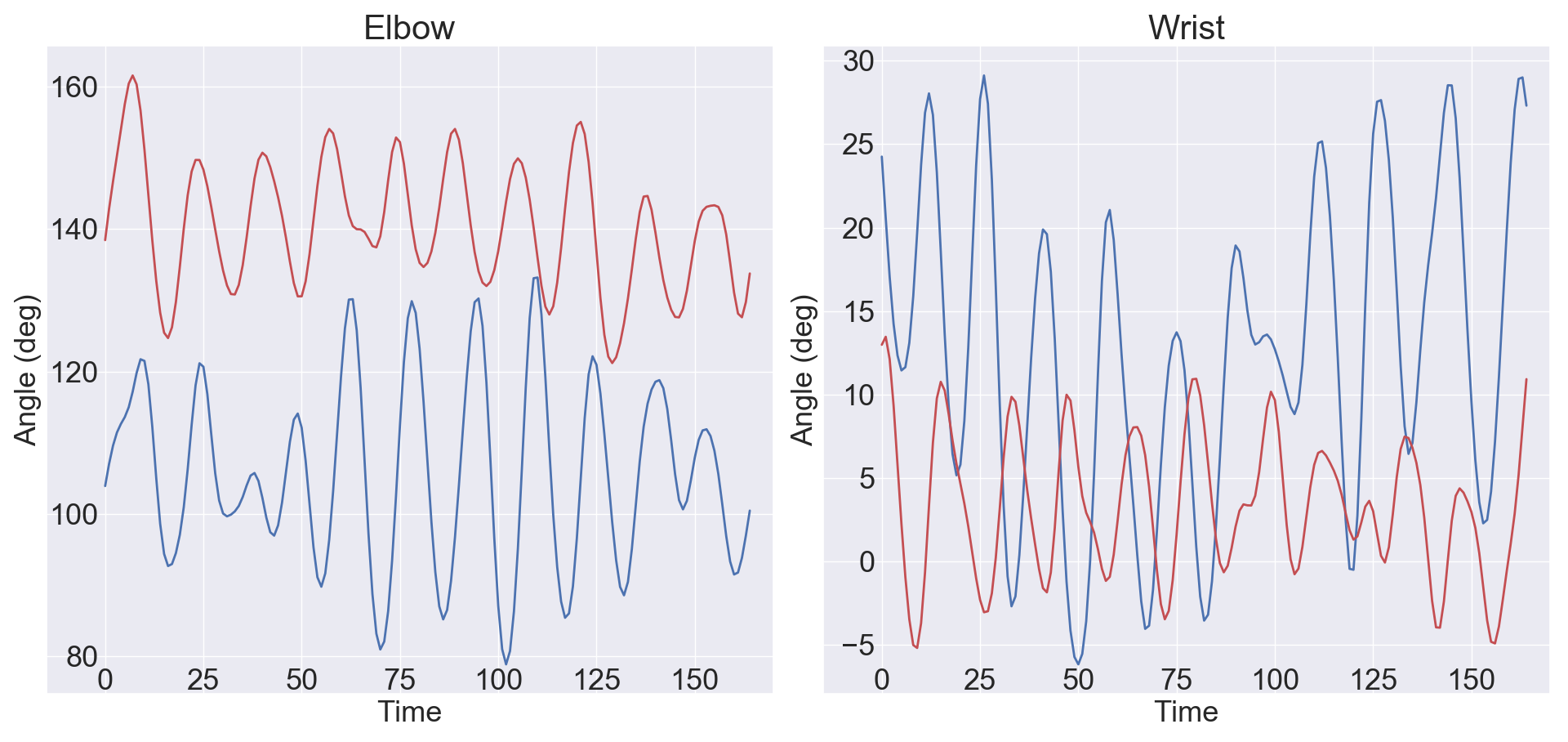}
	\caption{Trajectory samples of the same conditional distribution in red and blue. Modulation by conditioning generates similar motor patterns but different movements.}
	\Description{The 2007 Model D.}
	\label{fig:repros}
\end{figure}

\begin{figure}
	\centering
	\begin{subfigure}[b]{0.46\linewidth}
		\centering
		\includegraphics[width=\linewidth]{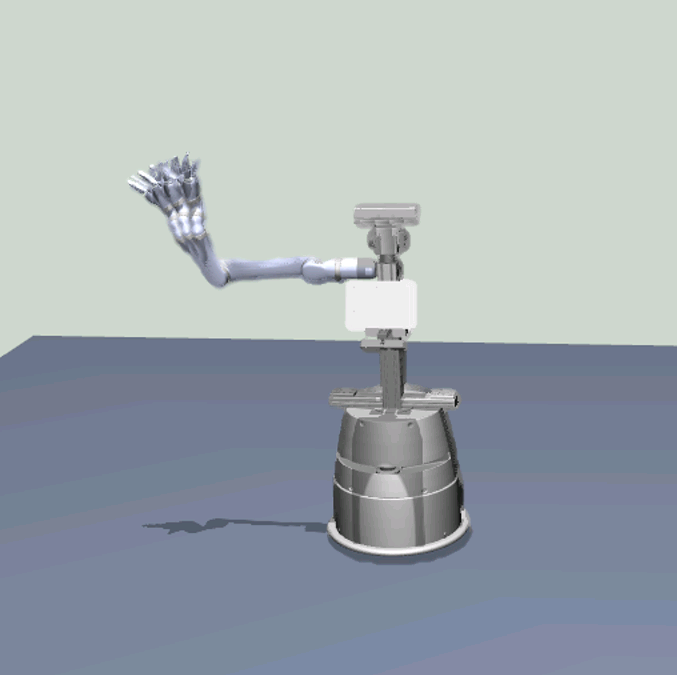}
	\end{subfigure}
	%\hfill
	\begin{subfigure}[b]{0.465\linewidth}
		\centering
		\includegraphics[width=\linewidth]{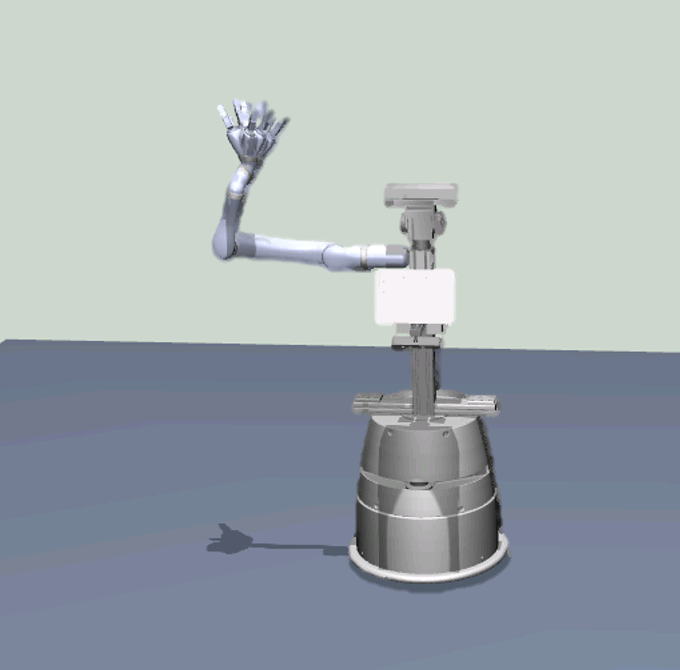}
	\end{subfigure}
	\caption{Overlaid images of the resulting waving patterns in simulated robot experiments. \textit{Left:} Conditioning on the elbow joint. \textit{Right:} Conditioning on the wrist joint.}
	\Description{The Model}
	\label{fig:Rcs-overlaid}
\end{figure}

We recorded 15 demonstrations of various postures and waving patterns with frequencies ranging between 1.5 and 3 Hz. The maximum number of oscillations per demonstration is 20 and thus we use K=25 conjugate basis functions for training. We use the movement data of the arm by taking into account D=5 degrees of freedom for training. We test the modulation capacity of the model by conditioning first on the random variable that corresponds to the amplitude of the 10th harmonic of the elbow joint and then on the 20th harmonic of the wrist joint. We observe the retrieved $\mu_{\bm{w}^r|\bm{w}^{r}_{k=10,elbow}=\,5}$ and $\mu_{\bm{w}^r|\bm{w}^{r}_{k=20,wrist}=\,4}$. The model is able to infer that a rhythmic movement of 10 elbow cycles should be accompanied by equivalent wrist cycles of similar amplitude and that for 20 wrist cycles the movement of the elbow should have low amplitude and higher frequency. Fig.{\ref{fig:spectrum}} shows the effect of modulation on the elbow and wrist joints in the frequency domain. Zero frequency terms are omitted for convenience of illustration. Fig.{\ref{fig:repros}} demonstrates the ability to preserve the stochastic variability of the generated movement in terms of posture, frequency, amplitude and phase during its modulation. Finally, we test the method on simulated robot experiments involving a robot with a manipulator of 6 degrees of freedom. The results are depicted in Fig.{\ref{fig:Rcs-overlaid}} and demonstrate the ability to execute various human-like wave gestures.   


\section{Conclusion}

Using the probabilistic movement primitives framework as a baseline we discussed a novel approach for the efficient learning and modulation of rhythmic movements through a probabilistic formulation. The proposed framework draws inspiration from the decomposition of signals as Fourier series, and extends ProMPs in the frequency domain by using complex exponential basis functions. The framework provides the structure needed in order to exploit the conditioning property for modulating rhythmic movements in an intuitive way, which was not possible in the original ProMPs framework. Albeit, these advantages come at the cost of abandoning the definition of a trajectory distribution which limits the modulation capacity of the framework to motions without a specific target in task space. Future work could include a user-based study of the naturalness of the generated gestures in comparison with existing motion generation methods in real robot experiments.
%
% The acknowledgments section is defined using the "acks" environment (and NOT an unnumbered section). This ensures
% the proper identification of the section in the article metadata, and the consistent spelling of the heading.
\begin{acks}
	This work was supported by Honda Research Institute Europe. 
\end{acks}

\bibliographystyle{ACM-Reference-Format}
\bibliography{HRI-sigconf}

\end{document}